\documentclass{article}

\usepackage{arxiv}

\usepackage[utf8]{inputenc} 
\usepackage[T1]{fontenc}    
\usepackage{hyperref}       
\usepackage{url}            
\usepackage{booktabs}       
\usepackage{amsfonts}       
\usepackage{nicefrac}       
\usepackage{microtype}      
\usepackage{wrapfig}        
\usepackage{multirow}       
\usepackage{xcolor}         
\usepackage{lipsum}         
\usepackage{graphicx}
\usepackage{natbib}
\usepackage{doi}
\usepackage{amsmath} 
\usepackage{cleveref}  
\newcommand{\yes}{$\checkmark$}

\newcommand{\method}{\textsc{LEAP-LLM}}
\newcommand{\mL}{\mathcal{L}}

\title{LEAP-LLM: Preserving Lead-Time Structure for Question-Conditioned ECG Interpretation}

\author{
    Aakash Agarwal \quad
    Yidan Shen \quad
    Xin Fu \quad
    Renjie Hu \\
    University of Houston
}

\renewcommand{\headeright}{Preprint}
\renewcommand{\shorttitle}{}

\begin{document}
\maketitle
\begin{abstract}
The electrocardiogram (ECG) records cardiac electrical activity across multiple leads and is used for detecting cardiac abnormalities. Recent ECG-language models enable large language models (LLMs) to interpret cardiac signals and answer clinically relevant questions. However, existing ECG-language models typically encode the ECG independently of the clinical question, forcing the ECG representation to retain broad and potentially irrelevant signal information. Moreover, these models often aggregate information across leads which can obscure lead-specific details needed for a particular question. To tackle these limitations, we introduce \method{}, an ECG-LLM architecture for \textbf{L}ead-aware \textbf{E}CG \textbf{A}lignment with \textbf{P}rompt conditioning. \method{} uses the clinical question to guide ECG feature extraction while preserving explicit lead and temporal structure. Task-trained instruction representations distill information from the question and guide two complementary branches. The Lead branch captures question-relevant temporal patterns within each ECG lead, while the Temporal branch summarizes relevant information across leads at each temporal position. Together, they provide the LLM with compact, question-relevant ECG information while preserving lead and temporal structure. \method{} reduces modality-token count by 89.0\% and prefill FLOPs by $7.2\times$ on average \textcolor{black}{relative to other ECG-LLMs baselines.}
Across four ECG benchmarks, \method{} achieves \textcolor{black}{better performance} , including a 3.4-percentage-point improvement in \textit{G12EC} over the strongest prior ECG-specific method. We further show that the model adaptively emphasizes the clinically relevant anatomical lead groups for different cardiac abnormalities, with lead importance patterns consistent with established diagnostic knowledge, providing interpretable evidence that the predictions \textcolor{black}{are supported by }clinically meaningful ECG structure.
\end{abstract}

\section{Introduction}

The electrocardiogram (ECG) is a widely used, non-invasive tool for detecting cardiac abnormalities such as arrhythmias, ischemia, and infarction \citep{namjouyan2025precision}. It captures cardiac electrical activity across multiple leads over time, with clinically relevant information distributed across both lead and temporal dimensions \citep{Berkaya2018ASO, hannun2019cardiologist}. Traditional deep learning methods have achieved strong performance on ECG analysis, but are typically developed for fixed prediction tasks such as abnormality classification \citep{dl1, dl2, dl3, dl4, dl5, dl6, dl7, dl8}.  Clinical ECG interpretation, however, is inherently more flexible meaning the same recording may be queried for rhythm, morphology, conduction abnormalities, etc. \citep{liu2024teach}. 

Large Language Models (LLMs) provide a natural interface for extending ECG models beyond fixed predictions toward question-conditioned interpretation.  Early ECG-LLM models explored ECG-text alignment for diagnosis and report generation \citep{cai2023jolt,zhao2025ecg}.  
More recent models such as PULSE ~\citep{liu2024teach} encodes
rendered ECG images and projects their features into visual tokens,
while GEM ~\citep{lan2025gem} projects signal and image features before combining them
with the textual query. These interfaces
provide question-independent ECG tokens, leaving the LLM to identify
the information relevant to each request during answer generation.

However, the evidence needed from an ECG depends on the question being asked. Different questions may need to refer to different leads and temporal regions within the same recording. For example, myocardial infarction (MI) is assessed through abnormalities in anatomically related lead groups, whereas rhythm assessment relies on temporal relationships between successive
beats~\citep{wagner2009aha,varma2021mhealth}. A  question-independent representation may therefore retain information irrelevant to the current question while discarding fine-grained details needed to answer it. This motivates using the question to guide evidence selection and aggregation when constructing the ECG tokens supplied to the LLM.
\begin{wrapfigure}{r}{0.28\textwidth}
    \centering
    \includegraphics[width=0.9\linewidth]{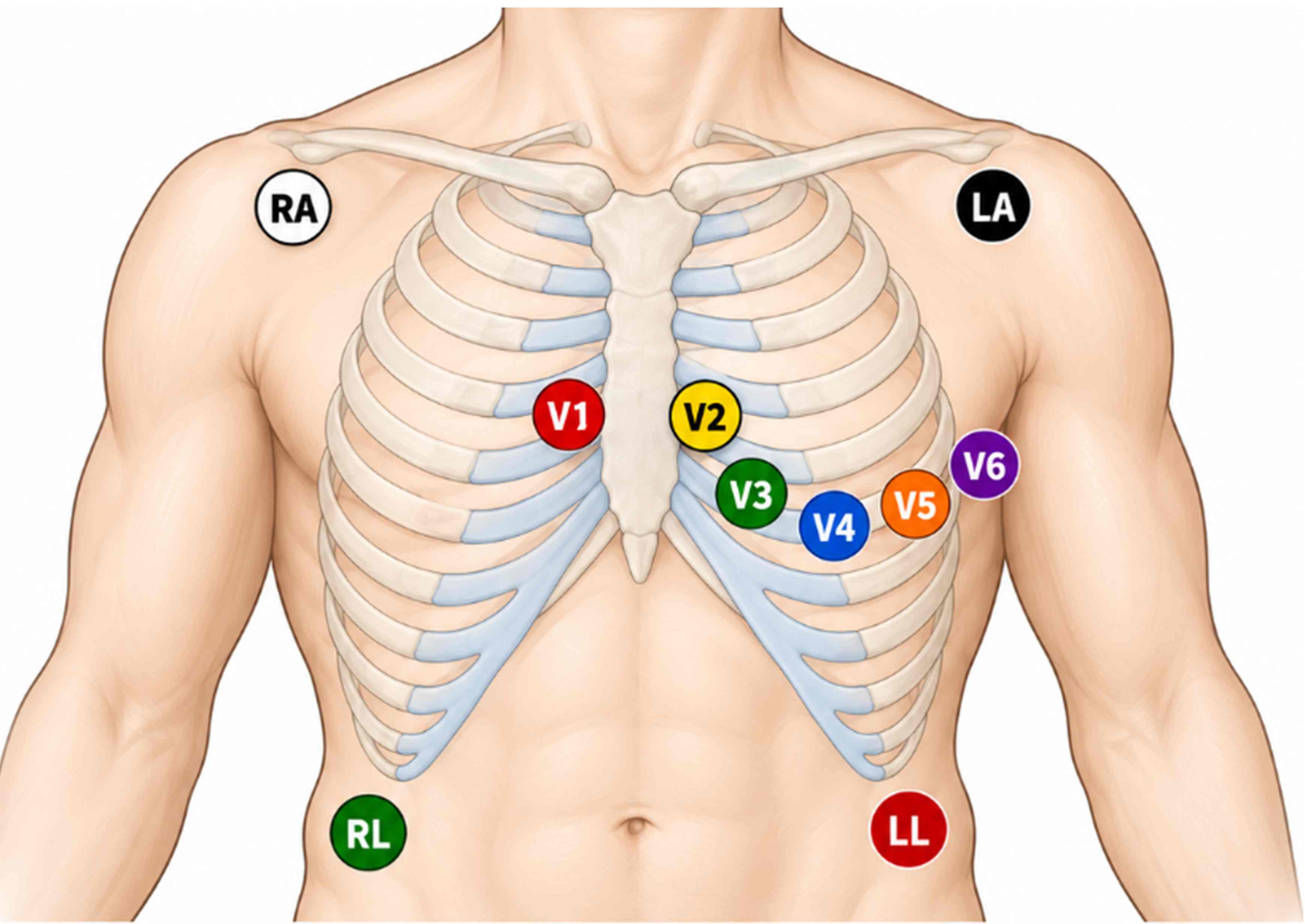}
\caption{\textbf{Electrode placement for 12-lead ECG acquisition.}
Ten electrodes generate the 12 ECG leads.}
    \label{fig:lead_positions}
\end{wrapfigure}

Several challenges remain in developing a question-guided ECG LLM.
{Firstly, global representation learning poses challenges for
retaining fine-grained details and lead identities.}
Existing ECG encoders do not explicitly preserve lead-specific waveform information in the representations provided to the LLM. Question-guided extraction requires representations that preserve these details and
explicitly identify their source leads and temporal
locations, allowing relevant evidence to be selected
while retaining its structural context.
{Secondly, question-ECG interaction poses challenges for efficient
information aggregation and structural preservation.} Relevant evidence
may involve temporal patterns within individual leads or comparisons
across leads at a given time. Simply flattening all ECG patches for
attention does not specify how the aggregated output should retain
these structures. The challenge is to produce compact ECG tokens that
preserve both lead-specific and time-indexed information.
{And Lastly, variation in question length introduces both
computational and semantic challenges.} Question length does not
necessarily reflect the amount of information relevant for ECG analysis. Clinical requests may be interspersed with redundant context or
instructions. Conditioning ECG features on the entire prompt at each interaction introduces unnecessary computation and can make signal aggregation sensitive to linguistically irrelevant tokens.

To address these challenges, we introduce \method{}, an ECG-LLM
framework for \emph{\textbf{L}ead-aware \textbf{E}CG \textbf{A}lignment with \textbf{P}rompt conditioning}.
We first use an encoder that maintains separate lead-specific features, with structural embeddings that make their lead identities and temporal locations explicit.
Building on these representations, we introduce question-conditioned dual-attention
alignment to aggregate question-relevant evidence without collapsing structural axes. The \textit{Lead Branch} aggregates temporal
evidence within each lead, whereas the \textit{Temporal Branch}
aggregates across leads at each temporal position, yielding lead-specific
and time-indexed representations, respectively. Both branches are guided
by task-trained instruction representations that distill
variable-length questions into a fixed set of task-oriented queries.
Together, this design
provides the LLM with compact, question-conditioned ECG tokens that
retain complementary lead and temporal organization.

We evaluate \method \  on PTB-XL, CODE-15\%, CSN, and G12EC, spanning multi-label abnormality detection and multiple-choice ECG interpretation. \method \ outperforms across all benchmarks, improving over prior ECG-specific methods by 0.7 percentage points on CODE-15\% AUC, 3.4 points on G12EC accuracy, and 0.4 points on CSN accuracy. 
At the same time, \method{} reduces the ECG modality-token count by 89.0\% and achieves a $7.2\times$ average fewer prefill FLOPs relative to comparable ECG-LLMs.

We further show that \method{} uses clinically meaningful
information preserved through explicit lead and temporal structure.
specifically, Lead-time attention analysis reveals selective focus on clinically
relevant temporal regions and anatomical lead groups, while targeted
lead-token interventions demonstrate the LLM's selective reliance on
anatomically relevant lead representations for diagnosis. Our main contributions are:

\begin{itemize}
    \item We introduce \method, a query-conditioned ECG-LLM framework that preserves explicit lead and temporal structure while extracting query-relevant signal information.
    
    \item We introduce complementary lead and temporal branches, guided by question-conditioned instruction tokens, to produce compact structured ECG representations for the LLM.
    
\item We achieve state-of-the-art performance on four ECG benchmarks while reducing the ECG modality-token count by 89.0\% and prefill FLOPs by $7.2\times$ on average relative to comparable ECG-LLMs.

    \item We establish the clinical relevance of the learned representations through attention and intervention analyses, showing temporally selective attention around salient cardiac events and selective reliance on anatomically relevant ECG leads.
    
\end{itemize}

\section{Related Work}

\subsection{Language Models for ECG Understanding}
Early work explored connecting ECG representations with language for diagnosis and report generation. 
\cite{yu2023zero} introduced retrieval-augmented approach for zero-shot ECG diagnosis, while JoLT~\citep{cai2023jolt} and MERL~\citep{merl} learned aligned ECG-text representations for language-guided prediction. 
ECG-CoCa and ECG-Chat~\citep{zhao2025ecg} further adapted vision-language architecture to directly process ECG time series input. ~\citet{wan2025meit} formulate ECG report generation as instruction tuning, transforming ECG-report pairs into conversational examples.

\textcolor{black}{
More recent ECG-specific multimodal language models substantially expand this paradigm. PULSE~\citep{liu2024teach} converts ECG signals into rendered images and instruction-tunes a multimodal LLM. GEM~\citep{lan2025gem} jointly incorporates raw ECG time series and 12-lead ECG images through separate encoders and cross-modal alignment, together with ECG-Grounding supervision. Other approaches operate directly on the ECG signal by converting it into discrete tokens. ECG-Byte~\citep{ecgbyte} adapts byte-pair encoding to tokenize ECG waveforms, while HeartLLM~\citep{heartllm} uses lead-wise encoding and quantization to construct ECG tokens for language-model processing. Beyond representation design, ECG-R1~\citep{ecgr1} combines signal and image inputs with reinforcement learning for grounded ECG reasoning. }

\color{black}
ESI \citep{esi} and FG-CLEP \citep{fgclep} explore ECG-text representation learning. ESI combines contrastive and captioning objectives, while FG-CLEP aligns local ECG segments with report tags to capture fine-grained waveform information. These approaches are complementary to \method{}. They improve the representations learned by the encoder, whereas \method{} uses the clinical question to select relevant information from these representations for the LLM. 
\color{black}

\textcolor{black}{These approaches primarily advance ECG-language modeling through representation alignment, multimodal integration, or reasoning supervision. In contrast, \method{} focuses on the {organization of the ECG representation exposed to the LLM}. Rather than representing the ECG as an image, a globally mixed feature sequence, or a single stream of discrete signal tokens, \method{} explicitly retains the physical lead and temporal axes, allowing the clinical question to selectively access lead-specific and temporally resolved evidence.}

\subsection{Time-Series-Language Alignment}

Beyond ECG, recent work has increasingly explored interactions between time series and language models. Several approaches incorporate textual information to improve conventional time-series prediction or classification  tasks~\citep{jinb,jinc,wang2024a,wang2024b}, while others use language models to generate descriptions or identify patterns from temporal signals~\citep{liu2024a}.  These methods are typically designed for generic time series, where channels are treated primarily as parallel signal dimensions. This formulation does not directly transfer to ECG, where each lead has a fixed physical identity and clinically meaningful relationship to the other leads.

More recent work has moved toward treating time series as a native modality for language-based reasoning and question answering. OpenTSLM~\citep{opentslm} evaluates natural-language reasoning across multivariate sensor domains, while ITFormer~\citep{itformer} introduces question-conditioned instruction tokens that guide channel and temporal aggregation for multimodal QA. \textcolor{black}{However, recent evaluations~\citep{artist, ecg_reasoning} highlight limitations of these approaches on ECG-specific question answering and reasoning tasks, motivating further investigation of ECG-specific representations.} \method{} builds on this broader direction but targets the structured nature of multi-lead ECGs. Rather than reducing ECG leads into a shared temporal representation, \method{} preserves complementary lead- and time-indexed representations, allowing the LLM to access question-relevant information while retaining explicit correspondence to physical ECG leads and temporal locations.

{\color{black}

\section{Method}
\label{sec:method}

We formulate ECG-language alignment as question-conditioned extraction over a structured lead-time representation. We design \method{} to preserve this structure while
selecting question-relevant information before it enters the LLM.
Given a multi-lead ECG $X$ and a clinical question $q$, our encoder first
extracts lead-specific temporal features and incorporates temporal,
lead-identity, and lead-category embeddings.
Task-trained instruction representations then distill the question into a
fixed set of queries for two complementary attention branches. The Lead
Branch retains lead-specific evidence, while the Temporal Branch retains
time-indexed evidence. Together, they form a compact, question-conditioned
ECG representation for answer generation, as illustrated in
Figure~\ref{fig:leap}.

\begin{figure}[t]
    \centering
    \includegraphics[width=\linewidth]{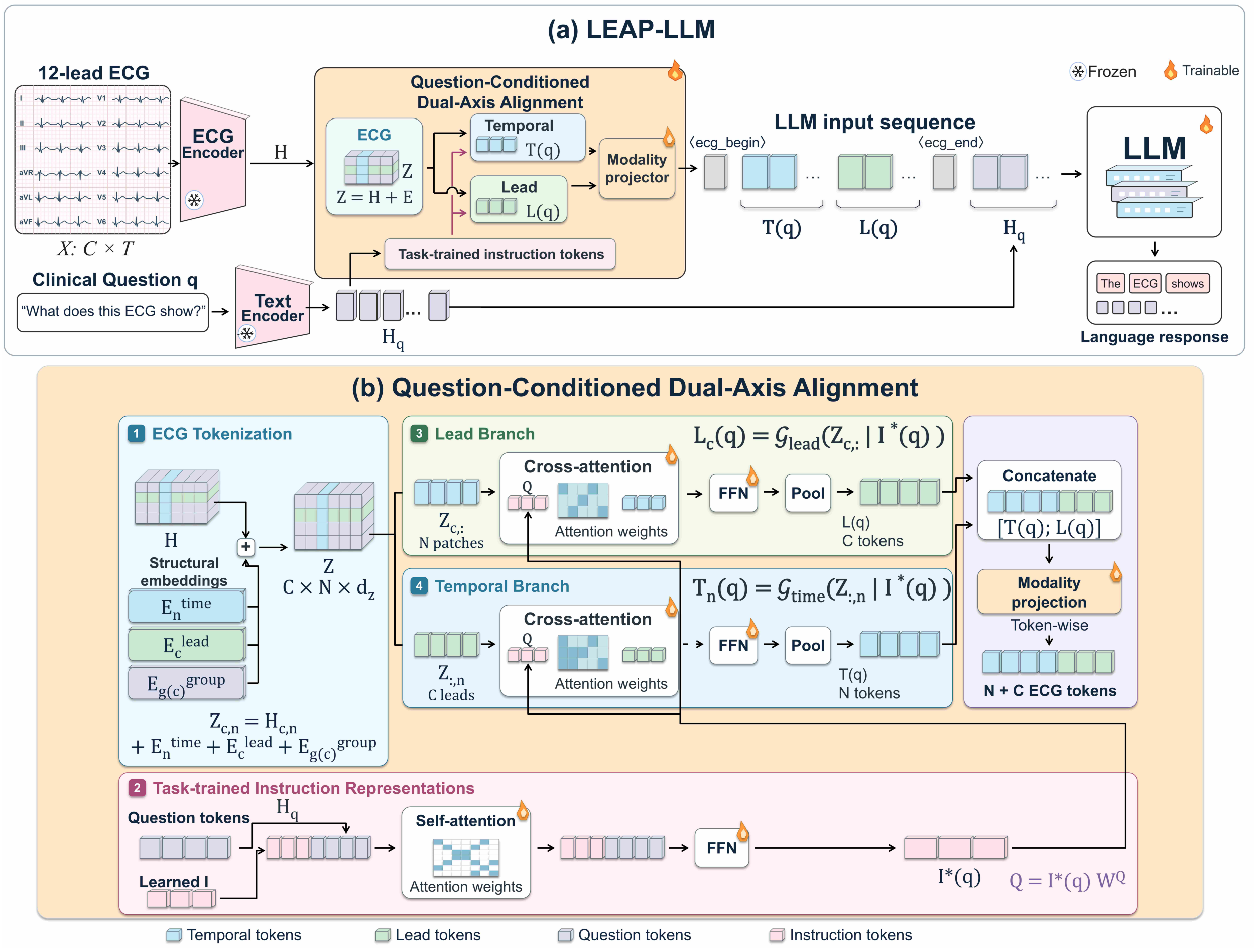}
    \caption{\textbf{Overview of \method{}.}
\textbf{(a)} A multi-lead ECG and clinical query are encoded separately. Task-trained instruction tokens condition a dual-axis alignment module to produce compact structured tokens for the LLM.
\textbf{(b)} The ECG is represented as lead-specific temporal tokens augmented with structural embeddings. Question-derived instruction tokens guide two complementary cross-attention branches: the \emph{Lead Branch}, which summarizes temporal evidence within each lead, and the \emph{Temporal Branch}, which summarizes cross-lead evidence at each temporal location. The resulting tokens are concatenated and projected into the LLM embedding space.}
    \label{fig:leap}
\end{figure}

\subsection{ECG Encoding with Structural Embeddings}
\label{sec:leap_structure}

Question-guided extraction requires lead-specific temporal features to
remain separately accessible. Our ECG encoder therefore maintains both
axes throughout signal encoding.
We represent the ECG signal as $X\in\mathbb{R}^{C\times T}$, where $C$
is the number of leads and $T$ is the number of samples per lead.
To preserve local temporal continuity, each lead is partitioned into overlapping patches of length $P$ and stride $S$, yielding $N=\lfloor(T-P)/S\rfloor+1$ temporal positions.
We construct an ECG encoder that processes each lead independently and
produces a feature tensor $H\in\mathbb{R}^{C\times N\times d_z}$ based on PatchTST~\citep{patchtst}.
Each feature $H_{c,n}$ represents a temporal patch from a specific lead,
allowing subsequent question-conditioned attention to access temporal
patterns within individual leads and compare features across leads.

\textbf{Structural embeddings.}
To make each feature's temporal location and lead information explicitly
available to the alignment module, we introduce three complementary
embeddings. A \emph{temporal-position embedding} $E^{\mathrm{time}}_n$
marks each patch's location within the recording, allowing the alignment
module to distinguish similar waveforms at different times.
A \emph{lead-identity embedding} $E^{\mathrm{lead}}_c$
identifies the source lead ~\citep{na2024stmem,kligfield2007standardization}, since the lead's electrical view affects how
its waveform is interpreted.
And finally to represent shared aspects of the lead measurement systems, we further
introduce a \emph{lead-category embedding} $E^{\mathrm{group}}_{g(c)}$.

The standard 12-lead ECG comprises three established lead families,
standard limb leads $\{\mathrm{I},\mathrm{II},\mathrm{III}\}$,
augmented limb leads $\{\mathrm{aVR},\mathrm{aVL},\mathrm{aVF}\}$,
and precordial leads $\{\mathrm{V1},\ldots,\mathrm{V6}\}$.
These families differ in their electrode combinations and reference
construction~\citep{kligfield2007standardization}.
Our category embedding follows this clinical classification, providing
a shared component within each lead family, while the lead-identity
embedding distinguishes individual leads.
The resulting representation is
\begin{equation}
    Z_{c,n}
    = H_{c,n}
    + E^{\mathrm{time}}_n
    + E^{\mathrm{lead}}_c
    + E^{\mathrm{group}}_{g(c)},
    \label{eq:leap_z}
\end{equation}
where $g(c)$ maps each lead to its category and
$Z\in\mathbb{R}^{C\times N\times d_z}$ retains separate lead and
temporal indices for subsequent aggregation.

\subsection{Task-Trained Instruction Representations}
\label{sec:leap_instruction}

Querying the ECG with all question tokens can introduce redundant linguistic context and makes cross-modal attention cost dependent on the variable or padded length of the question sequence. We therefore learn fixed-length instruction representations to distill task-relevant semantics and provide
a uniform query budget for both branches.

Let $H_q\in\mathbb{R}^{L_q\times d_I}$ denote the question embeddings.
We contextualize $M$ learnable instruction tokens
$I\in\mathbb{R}^{M\times d_I}$, with the question, through a self-attention
and feed-forward module $F_\phi$:
\begin{equation}
    I^*(q)
    = \left[F_\phi\bigl([I;H_q]\bigr)\right]_{1:M}
    \in\mathbb{R}^{M\times d_I}.
    \label{eq:leap_instruction}
\end{equation}
The instruction positions can attend to the entire question, and their
contextualized outputs form the queries used by both ECG branches.
Both $I$ and $F_\phi$ are optimized through the answer-generation objective
in Section~\ref{sec:leap_training}. Thus, the representation is trained
for extracting signal evidence that supports the answer, rather than
simply summarizing the question. This task supervision
encourages the contextualizer to emphasize relevant semantics and
downweight peripheral linguistic content.

Fixing $M$ independently of $L_q$ provides a uniform query length for
subsequent cross-modal attention. 
The full question is provided directly to the LLM, allowing the instruction representations to focus on ECG evidence extraction while preserving task semantics and response requirements.

\subsection{Question-Conditioned Dual-Attention Alignment}
\label{sec:leap_axes}

The structured feature grid and task-trained queries provide the two
inputs needed for question-conditioned evidence extraction. 
To extract question-relevant evidence while preserving lead and temporal
structure, we use parallel Lead and Temporal branches, each aggregating
over one axis while retaining the other in its output.

In each branch, the instruction representations guide the aggregation
of ECG features through multi-head cross-attention, with $I^*(q)$ as
queries and an ECG slice $S$ as keys and values. For branch
$b\in\{\mathrm{lead},\mathrm{time}\}$, we express the resulting
aggregation as
\begin{equation}
    \mathcal{G}_b(S \mid I^*(q))
    = \frac{1}{M}\sum_{m=1}^{M}
      \left[
      \mathcal{F}_b\!\left(
      \operatorname{CA}_b\bigl(I^*(q),S, S\bigr)
      \right)
      \right]_{m,:},
    \label{eq:leap_aggregation}
\end{equation}
where $\operatorname{CA}_b$ and $\mathcal{F}_b$ denote branch-specific
cross-attention and feed-forward transformations. The branches share
the instruction representations but have independent parameters.
Mean pooling is applied only over the $M$ instruction outputs, yielding
one $d_a$-dimensional representation for each ECG slice. Residual
connections and normalization are left implicit for clarity.

\textbf{Lead Branch: temporal evidence within each lead.}
For lead $c$, the branch accesses its complete encoded temporal sequence
$Z_{c,:}\in\mathbb{R}^{N\times d_z}$ and produces
$L_c(q)=\mathcal{G}_{\mathrm{lead}}(Z_{c,:}\mid I^*(q))$.
Each attention head assigns question-dependent weights to the temporal
patches through an $M\times N$ attention map. The resulting representation
summarizes relevant temporal evidence within that lead without combining
it with other leads. Applying this operation separately to all leads
produces $\mathbf{L}(q)\in\mathbb{R}^{C\times d_a}$, preserving one
question-conditioned representation per lead.

\textbf{Temporal Branch: cross-lead evidence at each temporal position.}
To retain temporal organization alongside the lead-specific summaries,
the Temporal Branch accesses all leads at each encoded patch position,
$Z_{:,n}\in\mathbb{R}^{C\times d_z}$, and produces
$T_n(q)=\mathcal{G}_{\mathrm{time}}(Z_{:,n}\mid I^*(q))$.
Its $M\times C$ attention map weighs the leads according to the question,
aggregating cross-lead evidence while retaining the temporal index $n$.
Repeating this operation across the recording produces
$\mathbf{T}(q)\in\mathbb{R}^{N\times d_a}$, a time-indexed sequence
that complements the lead-specific representations.

\textbf{Structured ECG tokens for the LLM.}
We concatenate the temporal and lead representations along the token
dimension and apply a learned projector to map each token into the LLM
embedding space. This tokenwise mapping preserves the correspondence to
individual leads and temporal positions without further pooling. The
combined sequence thus provides complementary views of the ECG:
lead-specific summaries of temporal evidence and time-indexed summaries
of cross-lead evidence. These question-conditioned ECG tokens are supplied
to the LLM together with the original question.

\subsection{Training Procedure}
\label{sec:leap_training}

\paragraph{Stage 0: ECG encoder pretraining.}
We first pretrain the ECG encoder through self-supervised masked signal
reconstruction. Each lead is encoded independently, and a lightweight
reconstruction head maps the encoded features back to the waveform
space. Reconstruction is supervised by the original waveform patches
using mean squared error evaluated only at masked positions. This
provides waveform-level supervision for the lead-specific representations
before multimodal alignment. After pretraining, we \textcolor{black}{discard the
reconstruction head } and keep the ECG encoder frozen throughout
Stages I and II. Pretraining details are provided in
Appendix~\ref{app:encoder_pretraining}.

\textbf{Stage I: ECG-language alignment.}
We freeze the pretrained ECG encoder and the LLM and train the
ECG-LLM interface, including the structural
embeddings, the instruction tokens and
contextualizer, the dual-attention branches, and the modality projector.
Let $\theta$ denote the interface parameters and $\psi$ the LLM
parameters. Given training examples $(X,q,a)\sim\mathcal{D}$, we
optimize the answer-generation objective
\begin{equation}
    \mathcal{L}_{\mathrm{ans}}(\theta,\psi)
    = -\mathbb{E}_{(X,q,a)\sim\mathcal{D}}
      \sum_{r=1}^{|a|}
      \log p_{\theta,\psi}
      \bigl(a_r\mid a_{<r},X,q\bigr).
    \label{eq:leap_loss}
\end{equation}
With $\psi$ fixed, the answer loss updates the entire interface, jointly
training the instruction representations and ECG aggregation modules along with the projector. This stage learns to extract question-relevant ECG evidence and present it for answer generation by the pretrained LLM.

\textbf{Stage II: ECG-language fine-tuning.}
Starting from the Stage-I checkpoint, we jointly fine-tune all LLM
parameters and all interface components, while keeping the ECG encoder
frozen. We retain the same objective in
~\eqref{eq:leap_loss}. Unlike the preceding stage, which adapts the interface to a fixed language model, this full-parameter fine-tuning allows the LLM to adapt to the complementary lead-specific and time-indexed ECG representations while continuing to refine question-conditioned evidence extraction. Additional training details are provided in Appendix~\ref{app:implementation}.}

\section{Experiments}

\subsection{Experimental Setup}

\textbf{Datasets and Evaluation.}
We train \method{} on ECGInstruct~\citep{liu2024teach} and
ECG-Grounding~\citep{lan2025gem}, comprising 1.16M and 30K
instruction-response pairs, respectively. ECGInstruct spans ECG feature
recognition, rhythm analysis and pathology identification across multiple question formats, while ECG-Grounding provides supervision
linking diagnostic interpretations to fine-grained ECG evidence. We evaluate
on four ECG-Bench~\citep{liu2024teach} datasets, PTB-XL and CODE-15\% for
multi-label abnormality detection, and G12EC and CSN for multiple-choice
abnormality detection. Following ECG-Bench, we report macro AUC, F1,
and Hamming Loss (HL) for multi-label tasks, and accuracy for multiple-choice
tasks. All downstream validation and test records are excluded from both ECG
encoder pretraining and ECG-language training. We additionally evaluate the
diagnostic information retained in the multimodal representations through
linear probing on PTB-XL and CODE-15 
provided in Appendix~\ref{apps:linearprobe}.

\textbf{Baselines.}
We compare \method{} against both general-purpose  and
ECG-specialized language models. General-purpose baselines include
InternVL~\citep{chen2024internvl}, Qwen-VL variants
\citep{wang2024qwen2vl,bai2025qwen25vl}, GPT-4o
\citep{openai2024gpt4o}, and Claude-3.5-Sonnet
\citep{anthropic2024claude}. ECG-specific baselines include
ECG-GPT~\citep{ecggpt}, HeartLLM~\citep{heartllm},
PULSE~\citep{liu2024teach}, GEM~\citep{lan2025gem}, and
ECG-R1~\citep{ecgr1}.

\textbf{Implementation Details.}
ECGs are resampled to 500\,Hz and represented as 10-second, 12-lead signals.
We use a frozen PatchTST-style ECG encoder with hidden dimension 512, producing
99 temporal patches per lead, and 128 question-conditioned instruction tokens.
The proposed interface compresses each ECG into 111 modality tokens before
passing them to the language model.  We evaluate a LLaVA based (Vicuna-7B) and a Qwen3-VL-8B backbone. Training follows the two-stage procedure in
Section~\ref{sec:method}. Stage~I optimizes only the ECG-LLM interface, while
Stage~II jointly fine-tunes the interface and language model with the ECG
encoder frozen. Models are trained on four NVIDIA L40S GPUs. Full
implementation and pretraining details are provided in
Appendix~\ref{app:implementation}.

\subsection{Results}
\begin{table}[!t]
\centering
\caption{ECG-Bench abnormality detection results. Best results are shown in \textbf{bold}, and second-best results are \underline{underlined}.}
\label{tab:ecgbench_results}
\small
\setlength{\tabcolsep}{6pt}
\renewcommand{\arraystretch}{1.05}

\begin{tabular}{l ccc ccc c c}
\toprule
\multirow{2}{*}{\textbf{Model}}
& \multicolumn{3}{c}{\textbf{PTB-XL}}
& \multicolumn{3}{c}{\textbf{CODE-15\%}}
& \textbf{G12EC}
& \textbf{CSN} \\
\cmidrule(lr){2-4}
\cmidrule(lr){5-7}
& F1 $\uparrow$
& AUC $\uparrow$
& HL $\downarrow$
& F1 $\uparrow$
& AUC $\uparrow$
& HL $\downarrow$
& Acc. $\uparrow$
& Acc. $\uparrow$ \\
\midrule

Random
& 33.2 & 50.3 & 50.1
& 15.0 & 48.8 & 32.1
& 12.1 & 11.6 \\

\midrule
\multicolumn{9}{c}{\textit{General-purpose MLLMs}} \\
\midrule

InternVL2-8B
& 14.3 & 50.6 & 27.8
& 16.1 & 55.8 & 17.7
& 37.5 & 47.7 \\

Qwen2-VL-7B
& 22.4 & 51.3 & 30.8
& 24.8 & 60.7 & 20.5
& 32.9 & 25.5 \\

Qwen2.5-VL-7B Instruct
& 16.8 & 51.5 & 27.6
& 26.2 & 57.9 & 19.5
& 13.8 & 12.2 \\

\midrule
\multicolumn{9}{c}{\textit{Proprietary MLLMs}} \\
\midrule

GPT-4o
& 28.3 & 55.6 & 26.2
& 24.9 & 59.9 & 15.7
& 49.2 & 57.5 \\

Claude-3.5-Sonnet
& 27.5 & 54.0 & 29.6
& 20.3 & 58.3 & 17.8
& 51.4 & 51.5 \\

\midrule
\multicolumn{9}{c}{\textit{ECG-specialized Methods}} \\
\midrule

ECG-GPT
& 53.9 & 69.5 & 20.1
& 40.1 & 68.9 & 17.4
& N/A & N/A \\

HeartLLM 
& 46.1 & 65.2  & 19.5
& 65.1 & 79.0 & 8.0
&11.4 & 13.3 \\

PULSE 
& 74.8 & 82.4 & 11.0
& 85.4 & 90.7 & 5.0
& 78.2 & 85.2 \\

GEM
& 75.8 & 83.4 & 11.0
& 86.4 & 91.5 & 4.7
& 80.5 & 86.2 \\

ECG-R1 
& 73.7 & 81.7 & 11.4
& 86.7 & 91.4 & {4.6}
& 84.5 & 90.4 \\

LEAP-LLM$_{\text{LLaVA}}$ 
& \textbf{76.2}
& \textbf{83.6}
& \textbf{10.5}
& \underline{87.2}
& \textbf{92.2}
& \underline{4.5}
& \underline{85.2}
& \underline{90.6} \\

LEAP-LLM$_{\text{Qwen3}}$
& \underline{75.9}
& \underline{83.5}
& \underline{10.7}
& \textbf{87.3}
& \underline{91.7}
& \textbf{4.5}
& \textbf{87.9}
& \textbf{90.8} \\

\bottomrule
\end{tabular}
\end{table}

\noindent \textbf{Predictive Performance.}
Table~\ref{tab:ecgbench_results} summarizes performance across the four ECG-Bench datasets. \method{} consistently improves over prior ECG-specific models across both PTB-XL and CODE-15\% as well as the out-of-domain G12EC and CSN benchmarks. With the same Vicuna-7B backbone as GEM, the LLaVA-based variant achieves stronger results on both PTB-XL and CODE-15\%, indicating that the gains arise from the structured lead-time ECG interface. Importantly, these improvements are not confined to a single evaluation metric. \method{} improves F1 and AUC while also reducing HL on both multi-label benchmarks. The improvements extend to unseen datasets, where the Qwen3-based variant surpasses ECG-R1 on both G12EC and CSN, including a 3.4-percentage-point gain on G12EC. Both \method{} variants outperform the strongest prior ECG-specific results on these out-of-domain benchmarks, suggesting that the advantage is not specific to a particular backbone. Together, these results show that preserving lead-time structure provides consistent benefits across datasets, evaluation metrics, and language-model backbones. \textcolor{black}{We also compare \method{} with general-purpose time-series models, ITFormer and OpenTSLM, and observe stronger performance on the evaluated ECG interpretation tasks. Detailed results are provided in Appendix~\ref{app:tsllm_comparison}.}


\begin{table}[h]
\centering
\caption{Inference efficiency comparison with ECG-LLM baselines.}
\label{tab:inference_efficiency}
\small
\setlength{\tabcolsep}{3.5pt}

\begin{tabular}{llcccc}
\toprule
\textbf{Method}
& \textbf{LLM Backbone}
& \textbf{ECG Tokens} 
& \textbf{Prefill FLOPs (T)} 
& \textbf{Prefill Latency (ms)} 
& \textbf{TTFT (ms)} \\
\midrule

HeartLLM & LLaMA-3.2-3B & 313   & \textbf{2.74}$^\dagger$ & 108.0 & 138.0 \\
GEM      & Vicuna-7B     & 676   & 10.73 & 65.8  & 90.7  \\
PULSE    & Vicuna-7B     & 2,879 & 43.66 & 242.7 & 270.9 \\
ECG-R1   & Qwen3-8B      & 862   & 13.52 & 145.9 & 149.1 \\
\midrule

\textbf{\method}
& Vicuna-7B
& \textbf{111}
& 3.13
& \textbf{33.1}
& \textbf{43.6} \\

\bottomrule
\end{tabular}
\begin{minipage}{\linewidth}
\footnotesize
$^\dagger$Lower FLOPs mainly reflect the smaller backbone, despite using
2.8$\times$ more modality tokens than \method{}. 4-bit NF4 inference adds
dequantization overhead not captured by FLOPs, increasing latency.
\end{minipage}
\end{table}
\noindent\textbf{Inference Efficiency.}
Table~\ref{tab:inference_efficiency} shows that \method{} is substantially more efficient than existing ECG-LLM approaches. Its structured interface represents each ECG with only 111 modality tokens, reducing the amount of ECG information processed by the LLM while preserving strong downstream performance. This compact representation translates directly into lower inference cost, with $3.4$-$14.0\times$ fewer prefill FLOPs and $2.1$-$6.2\times$ lower time to first token than the compared baselines. \method{} also achieves the lowest wall-clock prefill latency, and notably attains a lower TTFT than HeartLLM despite using a substantially larger language-model backbone. These results show that preserving lead-time structure enables a more compact and computationally efficient ECG-LLM interface without sacrificing interpretation quality. Full profiling details are provided in Appendix~\ref{app:inference_complexity}.

\subsection{Ablation}
\color{black}
\textbf{Contribution of Individual \method{} component:}
To evaluate the contribution of each component in \method{}, we conduct an ablation of the three key components, 1) lead-time structure (S), 2) question conditioning (Q), and 3) positional embeddings (P), as shown in Table~\ref{tab:ablation}. 

Starting from the question-agnostic baseline without these components (Row a), adding lead-time structure (Row b) improves accuracy by 4.1 percentage points on CSN and 3.0 points on G12, supporting the benefit of maintaining separate lead and temporal representations. Introducing question conditioning (Row c) further improves accuracy by 3.1 and 3.8 points, respectively, demonstrating the benefit of question conditioning feature extraction. Finally, adding lead and time positional embeddings (Row d) produces the incremental gains of 5.2 and 7.2 points, highlighting the value of our structural positional embeddings. The full configuration achieves the highest accuracy on both datasets, with overall improvements of 12.4 points on CSN and 14.0 points on G12 over the baseline.

\begin{table}[h]
\centering
\small
\caption{Contribution of lead-time structure (S), question conditioning (Q), and positional embeddings (P). Accuracy (\%) and gains over the preceding row are reported.}
\label{tab:ablation}
\begin{tabular}{lcccll}
\toprule
Variant & S & Q & P & CSN Acc $\uparrow$ & G12 Acc $\uparrow$ \\
\midrule
(a)          &      &      &      & 78.2        & 71.2 \\
(b)          & \yes &      &      & 82.3 (+4.1) & 74.2 (+3.0) \\
(c)          & \yes & \yes &      & 85.4 (+3.1) & 78.0 (+3.8) \\
(d) LEAP-LLM & \yes & \yes & \yes & \textbf{90.6} (+5.2) & \textbf{85.16} (+7.16) \\
\bottomrule
\end{tabular}
\end{table}
\color{black}
\textbf{Contribution of Individual Branch:} We ablate each branch of the dual-axis alignment module in the \method{}+LLaVA variant to isolate its contribution. The \textit{Lead-only} variant removes the Temporal Branch, retaining question-conditioned representations for each physical lead but discarding explicit cross-lead representations at each temporal location. The \textit{Temporal-only} variant removes the Lead Branch, retaining cross-lead representations over time but discarding separate lead-specific representations. Both ablations substantially reduce performance across the four benchmarks (Table \ref{tab:branch_ablation}). Retaining both branches achieves 76.20 F1 on PTB-XL, 87.20 F1 on CODE-15, 90.60\% accuracy on CSN, and 85.16\% on G12EC. Together, these results show that the two branches capture complementary information and neither is sufficient in isolation.
\begin{table}[h]
\centering
\caption{Architectural ablation of \method{} LlaVA across ECG-Bench datasets.}
\label{tab:branch_ablation}
\small
\begin{tabular}{llccc}
\toprule
\textbf{Dataset} & \textbf{Metric} & \textbf{Lead Only} & \textbf{Temporal Only} & \textbf{Full} \\
\midrule
PTB-XL  & F1  & \underline{51.68} & 42.43 & \textbf{76.20} \\
CODE-15 & F1  & 65.08 & \underline{77.59} & \textbf{87.20} \\
CSN     & Acc. & 74.30 & \underline{82.25} & \textbf{90.60} \\
G12EC   & Acc. & 56.01 & \underline{64.52} & \textbf{85.16} \\
\bottomrule
\end{tabular}
\end{table}
\section{Structural and Functional Analysis}
We test whether \method{} meaningfully preserves and uses lead-time structure through four complementary analyses. First, we evaluate whether different myocardial infarction territories show greater attention to their clinically associated lead groups. Second, we evaluate temporal selectivity by aligning attention to detected R peaks and measuring how attention varies across cardiac phase. \textcolor{black}{Third, we examine whether changing the question for the same ECG shifts \method{}'s attention toward relevant leads.} Finally, we test functional dependence of the tokens by replacing selected patient-specific lead tokens with their dataset-mean representations and measuring the resulting change in the likelihood of the correct answer. Together, these analyses assess whether the learned representations are temporally localized, anatomically structured, and functionally used by the LLM.

\begin{wrapfigure}{rt}{0.45\columnwidth}
    \centering
    \includegraphics[width=\linewidth]{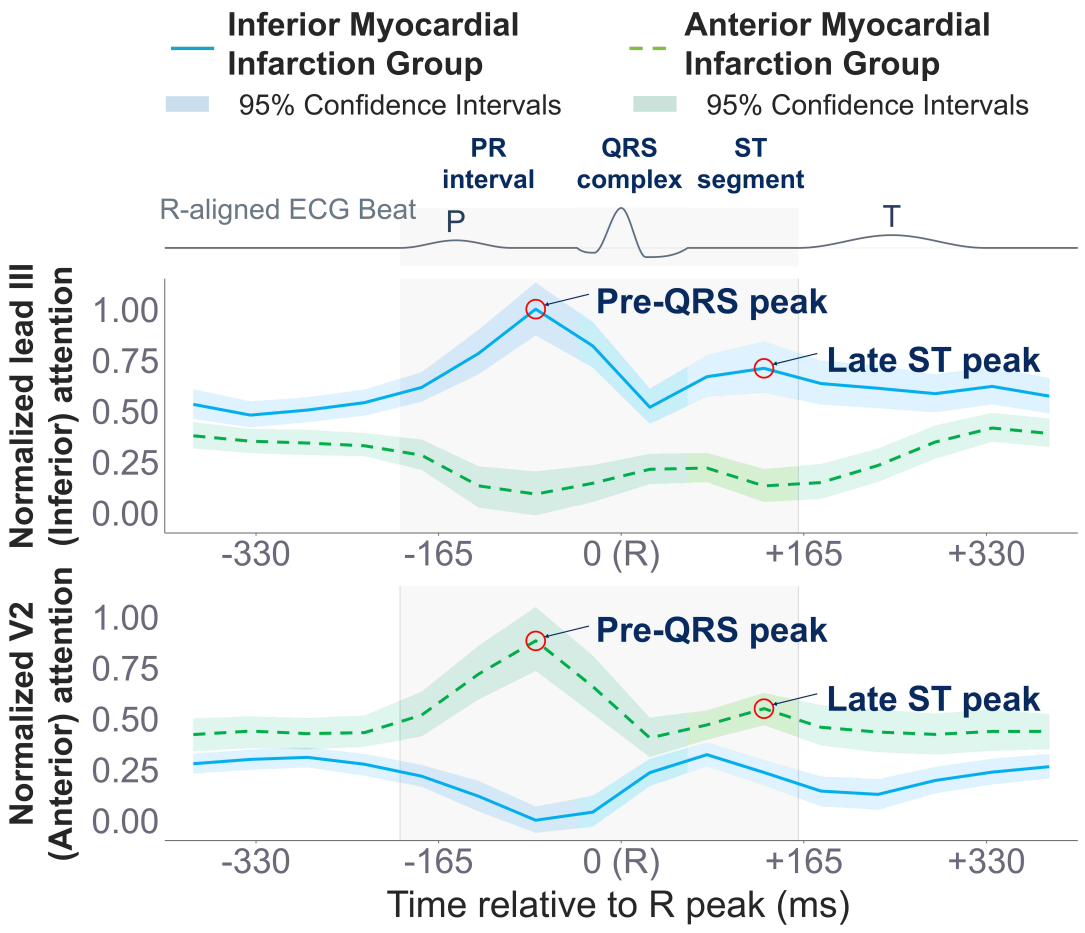}
\caption{\textbf{\method{} attention patterns in myocardial infarction groups.}
Inferior MI shows higher attention in inferior leads, while anterior MI shows higher attention in anterior leads. Attention peaks near clinically relevant QRS and ST-segment regions.}\label{fig:lead_phase_attention}
\end{wrapfigure}

Lead attention exhibits clear anatomical selectivity
(Figure~\ref{fig:lead_phase_attention}). Inferior MI patients receive
significantly greater attention than anterior MI patients over the inferior
leads (II, III, and aVF), whereas anterior MI patients receive greater attention than inferior MI patients over the anterior leads
(V1-V4). Thus, the lead-attention pattern changes with infarction territory and  emphasizes the lead group clinically associated with the
corresponding diagnosis \citep{localmi}.

\subsection{Lead-Temporal Selectivity.}

This selectivity is also localized in time. Both  groups show their strongest attention immediately before the QRS complex, with a secondary peak in the ST region. These regions correspond to waveform features used in myocardial infarction diagnosis and localization \citep{wagner2009aha}. Together, these patterns indicate that \method{} extracts diagnostic evidence from the appropriate anatomical leads at clinically relevant locations within the cardiac cycle. Details of R-peak alignment and attention calculation are provided in Appendix~\ref{app:attention_analysis}.

\color{black}
Additionally, when the same ECG is paired with a question about anterior rather than inferior MI, attention shifts from the inferior to the anterior lead group. Specifically, the inferior-lead attention share decreases from 79.8\% to 9.1\%, while the anterior-lead share increases from 7.7\% to 78.4\%. Because the ECG remains unchanged, this shift demonstrates that \method{} adapts its lead attention to the territory specified in the question. Details of the experiment are provided in Appendix~\ref{app:question_swap}.

\color{black}
\begin{wrapfigure}{r}{0.4\columnwidth}
\centering
\includegraphics[width=0.9\linewidth]{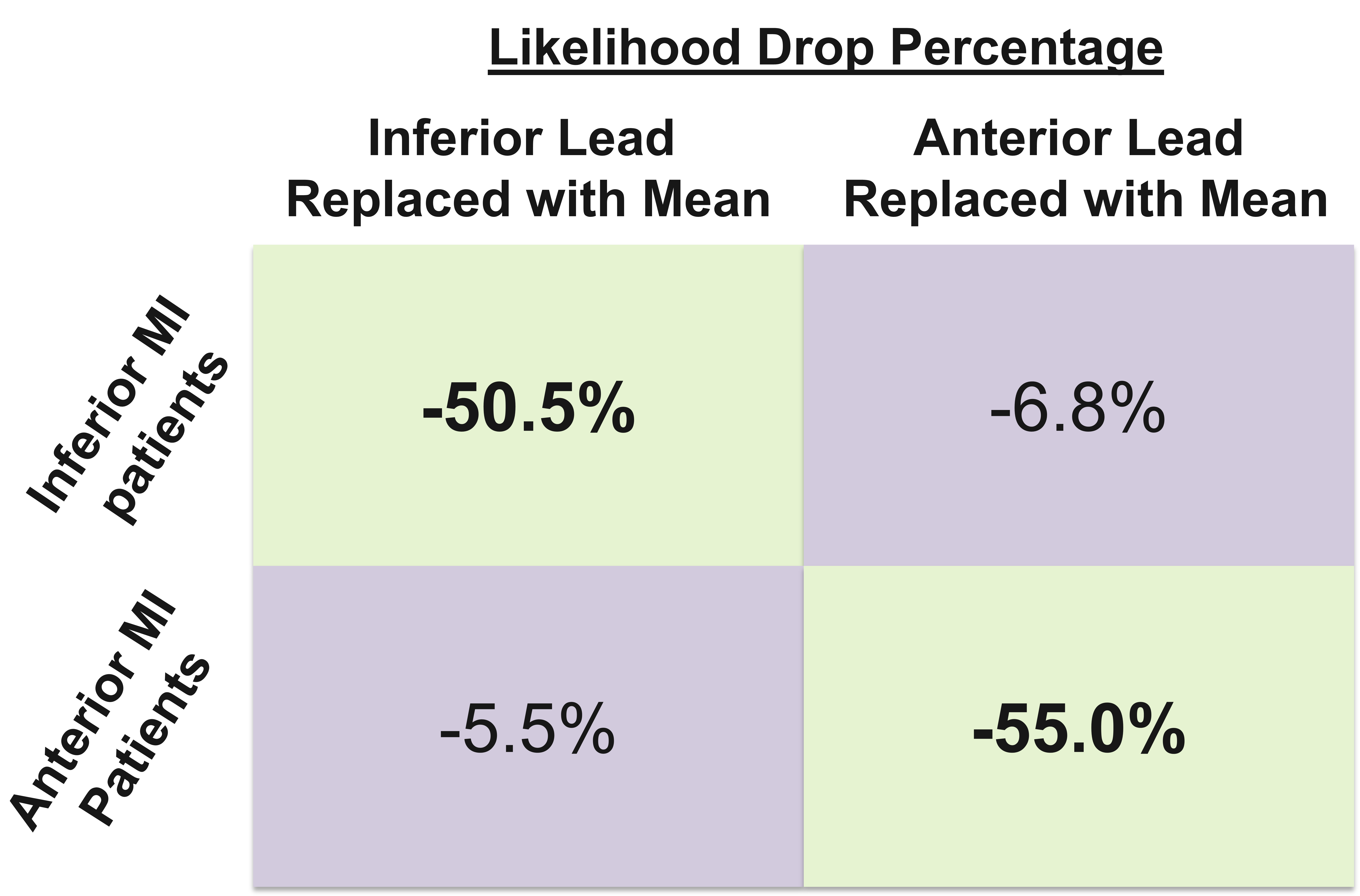}
\caption{ Reduction in the likelihood of the correct answer for different patients after replacing lead-specific tokens with their dataset-mean representations.}
\label{fig:lead_lesion}
\end{wrapfigure}

\subsection{Functional Dependence on Lead-Specific Representations}
Figure~\ref{fig:lead_lesion} provides functional evidence that the LLM selectively relies on lead representations associated with the relevant infarction territory.  We report the reduction in the likelihood of the correct response, where larger values indicate a stronger effect of replacing the corresponding lead tokens with their dataset-mean representations. For inferior MI samples, replacing the inferior leads reduces likelihood by 50.5\%, whereas replacing the anterior leads produces only a 6.8\% reduction. The pattern reverses for anterior MI, replacing the anterior leads causes a 55.0\% reduction, compared with only 5.5\% when the inferior leads are replaced. This crossed interaction provides functional evidence that the LLM selectively relies on anatomically relevant lead representations

Predictions for inferior and anterior MI are disproportionately affected when the lead group associated with the corresponding infarction territory is disrupted. Thus, preserving lead identity does not merely retain structural information, but determines which ECG evidence is used by the downstream language model.Further details on the intervention procedure are provided in Appendix~\ref{app:lead_lesion}.

\section{Conclusion}
We presented \method{}, a question-conditioned ECG-LLM interface designed to preserve the native lead and temporal organization of multi-lead ECG signals when translating them into language-model inputs. \method{} maintains complementary lead-specific and temporally resolved tokens while using the clinical question to selectively extract relevant signal evidence. Across multiple ECG benchmarks, this structured interface achieves strong performance while requiring substantially fewer modality tokens and lower inference cost than existing ECG-language approaches. Our structural analyses further show that the learned representations exhibit clinically meaningful organization across both lead and temporal dimensions, while targeted interventions demonstrate that the LLM selectively relies on anatomically relevant lead representations when forming its responses. Together, these findings suggest that preserving the intrinsic structure of physiological signals provides an effective and efficient interface for language-based reasoning over ECGs.

\section*{AI Disclosure}

In this work, we used generative AI tools to assist with manuscript writing and editing, including grammar correction, sentence restructuring, improving clarity and conciseness, and refining phrasing. The tools were not used to develop the research methodology, design experiments, generate or interpret experimental results, or formulate the scientific conclusions of this work. We reviewed all AI-assisted text and revised it as necessary. We take responsibility for the final content of this work, including all text, claims, and other material.

\newpage
\bibliographystyle{unsrtnat}
\bibliography{ref}

\appendix
\section{Electrocardiogram (ECG) Background}
\textbf{ECG Acquisition}
A standard 12-lead electrocardiogram (ECG) records cardiac electrical activity from multiple spatial viewpoints using ten physical electrodes, four limb electrodes and six precordial electrodes. The limb electrodes generate six limb leads, with the augmented leads derived from the standard limb leads as
\[
\mathrm{aVR}=-\frac{\mathrm{I}+\mathrm{II}}{2},\qquad
\mathrm{aVL}=\frac{\mathrm{I}-\mathrm{III}}{2},\qquad
\mathrm{aVF}=\frac{\mathrm{II}+\mathrm{III}}{2}.
\]
The six precordial electrodes provide leads
$\{\mathrm{V1},\mathrm{V2}, \mathrm{V3}, \mathrm{V4}, \mathrm{V5},\mathrm{V6}\}$, resulting in 12 ECG leads from 10 physical electrodes.

\textbf{ECG Temporal Structure and Diagnostic Information}
Many cardiac abnormalities are identified by presence of a waveform change at different temporal locations.. The P wave reflects atrial depolarization, the QRS complex primarily reflects ventricular depolarization, and the T wave reflects ventricular repolarization. Consequently, different abnormalities manifest through changes in distinct temporal regions or relationships between them. For example, rhythm abnormalities may be identified from beat-to-beat regularity, R-R intervals, and P-wave--QRS relationships, whereas conduction and morphological abnormalities may appear through changes in QRS duration, waveform shape, or ST-T segments~\citep{Berkaya2018ASO,varma2021mhealth}. Thus, preserving temporal location is important because diagnostically relevant evidence is distributed non-uniformly across the cardiac cycle.

\textbf{Lead-dependent diagnostic information.}
Many cardiac abnormalities are identified not only by the presence of a waveform change, but also by the leads in which that change occurs. This is particularly important for myocardial infarction (MI), where lead groups correspond approximately to different anatomical regions of the myocardium. Inferior MI is primarily associated with abnormalities in leads II, III, and aVF, whereas anterior infarction is reflected predominantly in the precordial leads V1-V4~\citep{wagner2009aha,localmi}. 

These properties motivate treating a multi-lead ECG as a structured
lead-time signal rather than as an unordered collection of features.
The temporal axis describes how cardiac electrical activity evolves within and
across beats, while the lead axis identifies the spatial viewpoint from which
that activity is measured. In \method{}, these two axes are explicitly
preserved so that a clinical question can selectively retrieve evidence from
the leads and temporal regions relevant to the requested interpretation.
\section{Implementation and Training Details}
\label{app:implementation}

\subsection{ECG Preprocessing}

All ECG recordings are resampled to 500\,Hz and represented as
10-second, 12-lead signals. Each input therefore contains
$C=12$ leads and $T=5000$ temporal samples. Signals are provided to
the ECG encoder in their original voltage scale. We do not apply
per-record standardization because absolute waveform amplitude
contains diagnostically relevant information \citep{voltage}.

Each lead is partitioned into overlapping patches of length $P=100$
samples with stride $S=50$, corresponding to a patch duration of
200\,ms and a displacement of 100\,ms between consecutive patches.
The resulting number of patches per lead is
\begin{equation}
N
=
\left\lfloor
\frac{T-P}{S}
\right\rfloor + 1
=
\left\lfloor
\frac{5000-100}{50}
\right\rfloor + 1
=
99.
\end{equation}
Thus, the ECG encoder receives 99 temporal patches from each of the
12 leads.

\subsection{ECG Encoder Pretraining}
\label{app:encoder_pretraining}

\textcolor{black}{We pretrain the PatchTST encoder on 805{,}926 ECG recordings, 17{,}104 from the PTB-XL v1.0.3 training split and 788{,}822 from the MIMIC-IV-ECG v1.0 training split  using a self-supervised masked signal reconstruction objective.} The temporal patches are embedded into a 512-dimensional latent space. During pretraining, 40\% of the input patches are randomly masked, and the encoder is trained to reconstruct the corresponding raw ECG signal from the remaining visible context. A lightweight reconstruction head maps the encoded representations back to the signal space, and the model is optimized using mean squared error computed over the masked regions:
\[
\mathcal{L}_{\mathrm{rec}}
=
\frac{1}{|\mathcal{M}|}
\sum_{i \in \mathcal{M}}
\left\|
\hat{\mathbf{x}}_i-\mathbf{x}_i
\right\|_2^2,
\]
where $\mathcal{M}$ denotes the set of masked patches, $\mathbf{x}_i$ is the original ECG patch, and $\hat{\mathbf{x}}_i$ is its reconstruction. This objective encourages the encoder to capture local waveform morphology as well as dependencies across longer temporal contexts without relying on diagnostic labels. After pretraining, the reconstruction head is discarded and the pretrained PatchTST encoder is used to initialize the ECG encoder in our multimodal framework.

\color{black}
\subsection{Encoder Validation}
To evaluate the diagnostic information captured by the pretrained encoder, we perform linear probing on held out CODE-15\% dataset. We randomly sample 600 single-label examinations from each of seven classes, yielding 4{,}200 examinations. An 80/20 examination-level split with seed~0 produces 3{,}360 training and 840 test examinations.

ECGs are resampled from 400~Hz to 500~Hz. The encoder remains frozen, and its features are averaged across leads and patches to obtain a 512-dimensional representation per examination. Features are standardized using training-set statistics. We then fit a one-vs-rest logistic regression classifier with $C=1$, the L-BFGS solver, and a maximum of 1{,}000 iterations. Only the classifier is trained on CODE-15\% labels

\begin{table}[h]
\centering
\small
\caption{Frozen-encoder linear-probe AUROC on the sampled CODE-15\% cohort of 840 test examinations.}
\label{tab:encoder_probe}
\begin{tabular}{lr}
\toprule
Class & AUROC $\uparrow$ \\
\midrule
Normal ECG            & 0.989 \\
First-degree AV block & 0.983 \\
RBBB                  & 0.988 \\
LBBB                  & 0.992 \\
Sinus bradycardia     & 0.995 \\
Sinus tachycardia     & 0.998 \\
Atrial fibrillation   & 0.991 \\
\midrule
Macro-AUROC           & \textbf{0.991} \\
\bottomrule
\end{tabular}
\end{table}

As shown in Table~\ref{tab:encoder_probe}, the frozen encoder achieves a macro-AUROC of 0.991, with per-class AUROCs ranging from 0.983 to 0.998. These results demonstrate that the pretrained representations contain diagnostic information accessible to a linear classifier on a dataset excluded from encoder pretraining.
\color{black}
\subsection{Model Configuration}

We use a 6-layer PatchTST-style encoder with a hidden dimension of
$d_{\mathrm{enc}}=512$ and 8 self-attention heads. The encoder processes
each lead independently. 

Our architecture contains 128 learnable instruction tokens. These
tokens are contextualized by the clinical question and shared between
two independently parameterized cross-attention branches. The Lead
Branch produces one representation for each of the 12 physical leads,
whereas the Temporal Branch produces one representation for each of the
99 temporal patches. Their outputs are concatenated to form
\begin{equation}
N_{\mathrm{ECG}} = C + N = 12 + 99 = 111
\end{equation}
structured ECG tokens. A learned modality projector maps these tokens from the interface dimension into the embedding dimension of the language model without altering the token axis.

We instantiate the proposed interface with the two language-model backbones reported in  Table~\ref{tab:ecgbench_results}: a LLaVA-based backbone for the first model variant and Qwen3-VL-8B for the second variant. More details about the training are presented in Table~\ref{tab:training_configuration}.

\begin{table}[t]
\centering
\caption{Optimization settings for ECG encoder pretraining and the two ECG-language training stages.}
\label{tab:training_configuration}
\small
\setlength{\tabcolsep}{4pt}
\renewcommand{\arraystretch}{1.05}
\begin{tabular}{llll}
\toprule
\textbf{Setting}
& \textbf{Stage 0}
& \textbf{Stage I}
& \textbf{Stage II} \\
& \textit{Encoder Pretraining}
& \textit{Alignment}
& \textit{Fine-tuning} \\
\midrule

ECG encoder
& Trainable
& Frozen
& Frozen \\

ECG-LLM interface
& -- 
& Trainable
& Trainable \\

Language model
& --
& Frozen
& Trainable \\

Training objective
& Masked reconstruction
& Answer generation
& Answer generation \\

Optimizer
& AdamW
& AdamW
& AdamW \\

Peak learning rate
& $1\times10^{-3}$
& $5\times10^{-4}$
& $1\times10^{-5}$ \\

Scheduler
& Cosine decay
& Cosine decay
& Cosine decay \\

Warm-up
& 5\%
& 5\%
& 5\% \\
Epochs
& 10
& 1
& 1 \\

Batch size
& 48
& 48
& 48 \\

Gradient accumulation
& 2
& 8
& 8 \\

Precision
& bf16
& bf16
& bf16 \\

\bottomrule
\end{tabular}
\end{table}

\section{Diagnostic Information in Multimodal Representations}
\label{apps:linearprobe}
\subsection{Linear Probing}

We use linear probing to evaluate how much diagnostically discriminative
information remains accessible in the multimodal representations supplied
to the language model. We probe the output of each model's \textcolor{black}{alignment} immediately before it is consumed by the LLM, rather than probing the underlying ECG or vision encoder. This isolates the quality of
the modality representation exposed to the language model.

\noindent\textbf{Representation extraction.}
For each model, we extract the modality representation immediately before it is provided to the LLM. For \method, this corresponds to the concatenated lead and temporal branch outputs after projection into the LLM embedding space, yielding 111 tokens: 12 lead tokens and 99 temporal tokens. For GEM, we concatenate the projected time-series and image tokens. For PULSE, we extract the projected visual tokens produced from the rendered ECG image.

 All three models expose 4096-dimensional representations at the LLM interface. We use the LLaVA-based variant of \method{} for this comparison so that \method{}, GEM, and PULSE share the same decoder family and interface dimensionality. All multimodal model parameters remain frozen during representation extraction and probe training.

\noindent\textbf{Evaluation cohorts.}
We conduct probing experiments on the official ECGBench PTB-XL and
CODE-15\% evaluation cohorts. The corresponding native ECG waveforms
are used for \method, while the official ECGBench-rendered images are
used for PULSE. GEM receives both representations of the same
examination. The evaluated records are held out from the instruction-
tuning data used to train the compared models.

The PTB-XL cohort consists of 2,082 records from the official test split. PTB-XL defines five diagnostic superclasses: normal ECG (NORM), myocardial infarction (MI), ST/T change (STTC), conduction disturbance (CD), and hypertrophy (HYP). Because these superclasses are not mutually
exclusive, we retain only examinations assigned to exactly one
superclass, providing an unambiguous target for this probing analysis.

For CODE-15\%, we evaluate seven abnormalities: normal ECG,
first-degree atrioventricular block (1dAVb), right bundle branch block
(RBBB), left bundle branch block (LBBB), sinus bradycardia (SB), sinus
tachycardia (ST), and atrial fibrillation (AF). We retain the
multi-label formulation and independently probe each abnormality using
a one-vs-rest target.

\noindent\textbf{Probe training.}
The extracted representations are frozen before probe fitting. We train an
independent one-vs-rest logistic regression classifier for every
diagnostic category. This gives five classifiers per model for PTB-XL
and seven classifiers per model for CODE-15\%.

For each dataset, examinations are randomly divided into 80\% training
and 20\% testing partitions using seed 0. For PTB-XL, this results in
1,274 training and 319 testing examinations. The identical
examination-level partition is used for \method, GEM, and PULSE,
ensuring that differences cannot be attributed to the probe split.

We use scikit-learn's \texttt{LogisticRegression} implementation with
the \texttt{lbfgs} optimizer and \texttt{max\_iter=1000}. All other
hyperparameters use the default scikit-learn configuration. Performance
is measured using the area under the receiver operating characteristic
curve (ROC-AUC) for each one-vs-rest classifier. 
\begin{table}[t]
\centering
\caption{
Per-class linear probing ROC-AUC on PTB-XL and CODE-15\%.
The best result for each class is shown in bold.
}
\label{tab:probing_per_class}
\small
\setlength{\tabcolsep}{6pt}
\renewcommand{\arraystretch}{1.05}
\begin{tabular}{llccc}
\toprule
\textbf{Dataset}
& \textbf{Class}
& \textbf{\method}$_\text{LlaVA}$
& \textbf{GEM}
& \textbf{PULSE} \\
\midrule

\multirow{6}{*}{PTB-XL}
& NORM & \textbf{0.8914} & 0.8199 & 0.7614 \\
& MI   & \textbf{0.9290} & 0.8313 & 0.6415 \\
& STTC & \textbf{0.9377} & 0.8735 & 0.8404 \\
& CD   & \textbf{0.8657} & 0.8095 & 0.7955 \\
& HYP  & \textbf{0.8515} & 0.7701 & 0.7677 \\
& \textbf{Macro}
& \textbf{0.8951} & 0.8208 & 0.7613 \\
\midrule

\multirow{8}{*}{CODE-15\%}
& Normal ECG & \textbf{0.9918} & 0.9108 & 0.8756 \\
& 1dAVb      & \textbf{0.9663} & 0.9316 & 0.8415 \\
& RBBB       & \textbf{0.9877} & 0.9824 & 0.9347 \\
& LBBB       & \textbf{0.9981} & \textbf{0.9981} & 0.9892 \\
& SB         & \textbf{0.9935} & 0.9668 & 0.9711 \\
& ST         & \textbf{0.9952} & 0.9813 & 0.9621 \\
& AF         & \textbf{1.0000} & 0.9659 & 0.9326 \\
& \textbf{Macro}
& \textbf{0.9904} & 0.9624 & 0.9295 \\
\bottomrule
\end{tabular}
\end{table}

\noindent\textbf{Results.}
As shown in Table~\ref{tab:probing_per_class}, \method{} produces the
strongest linearly separable representations on both datasets. On
PTB-XL, \method{} achieves an AUC of 0.8951, improving upon GEM and
PULSE-7B by 0.0743 and 0.1338, respectively. The improvement is
observed across all five diagnostic superclasses, including
morphology-oriented categories such as MI, STTC, and HYP.

On CODE-15\%, \method{} achieves a macro-AUC of 0.9904, compared with
0.9624 for GEM and 0.9295 for PULSE. \method{} obtains the highest or
tied-highest AUC for all seven abnormalities. These results indicate that our model consistently preserves diagnostically relevant information across diverse abnormalities, demonstrating the benefit of maintaining explicit lead and temporal structure at the ECG-LLM interface.

\color{black}
\section{Comparison with General-Purpose Time-Series Models}
\label{app:tsllm_comparison}

We compare \method{} with ITFormer \cite{itformer} and OpenTSLM \cite{opentslm}. All models are evaluated on the same PTB-XL and Code-15\% test split. Since original ITFormer is trained on the same instruction-tuning data as \method{}, whereas OpenTSLM is evaluated using its released ECG checkpoint without additional fine-tuning.

\textbf{ITFormer.}
ITFormer was originally developed and trained on Engine MT-QA, an aero-engine time-series question-answering dataset \cite{itformer}. We therefore retrain it on ECG data using the authors' implementation with Qwen2.5-3B-Instruct. Specifically, we use the same 356{,}423 instruction-tuning examples as \method{}-3B, drawn from the PTB-XL, ECG-QA, and CODE-15\% training splits. Each ECG is resampled to 600 samples per lead and standardized per lead, and the examples are converted to ITFormer's question-answer format. Following its training procedure, we first pretrain the time-series encoder on the training ECGs and then train the interface with the encoder and LLM frozen.

\textbf{OpenTSLM.}
Since OpenTSLM released \texttt{llama-3.2-3b-ecg-sp} checkpoint has already trained on ECG-specific dataset (ECG-QA) we do not retrain it. Following its input format, we provide the 12 leads as separate time series sampled at 100~Hz and use its prompt template with greedy decoding. 

\begin{table}[h]
\centering
\small
\setlength{\tabcolsep}{4pt}
\caption{Comparison with ITFormer and OpenTSLM at the 3B scale. All metrics are reported as percentages. HL denotes Hamming loss.}
\label{tab:tsllm_comparison}
\begin{tabular}{lrrrrrr}
\toprule
 & \multicolumn{3}{c}{PTB-XL} & \multicolumn{3}{c}{CODE-15\%} \\
\cmidrule(lr){2-4} \cmidrule(lr){5-7}
Model & F1 $\uparrow$ & AUROC $\uparrow$ & HL $\downarrow$ & F1 $\uparrow$ & AUROC $\uparrow$ & HL $\downarrow$ \\
\midrule
ITFormer-3B  & 12.3 & 50.0 & 28.1 & 26.5 & 61.7 & 21.5 \\
OpenTSLM-3B  & 21.6 & 50.9 & 30.7 & 14.5 & 52.9 & 16.3 \\
\method{}-3B & \textbf{71.2} & \textbf{80.4} & \textbf{12.7} & \textbf{83.5} & \textbf{88.7} & \textbf{5.8} \\
\bottomrule
\end{tabular}
\end{table}

\paragraph{Results.}
As shown in Table~\ref{tab:tsllm_comparison}, \method{}-3B  achieves the highest F1 and AUROC and the lowest Hamming loss on both datasets. With matched instruction-tuning data, it exceeds ITFormer by 30.4 AUROC points on PTB-XL and 27.0 points on CODE-15\%. The released OpenTSLM checkpoint also shows limited AUROCs of 50.9 and 52.9, respectively.

\textbf{Note on General Purpose Time Series Models:}
Independent studies also report limited performance for these models on ECG tasks. ARTIST \cite{artist} fine-tunes 4B versions of ITFormer and OpenTSLM on its own instruction-tuning data and reports ECG-QA F1 scores of 49.91 and 41.00, respectively, compared with a random-guess F1 of 50.00. ECG-Reasoning-Benchmark \cite{ecg_reasoning} reports initial-diagnosis accuracy for OpenTSLM-3B of 54.8\% on PTB-XL and 51.6\% on MIMIC-IV-ECG, whereas PULSE, GEM, and ECG-R1 achieve 80.9-85.4\% and 72.1-80.2\%, respectively. 

These independent findings align with our results and suggest that general-purpose time-series LLMs are less well suited to ECG interpretation than ECG-specific models.

\color{black}
\section{Inference Efficiency Measurement}
\label{app:inference_complexity}

We evaluate the inference efficiency of \method{} and the ECG-LLM
baselines under a unified profiling protocol. All methods are evaluated
using the same input example, prompt, hardware configuration, batch size,
and decoding length. We report modality-token count, prefill FLOPs,
prefill latency, time to first token (TTFT), end-to-end generation latency,
and peak GPU memory.

\subsection{Evaluation Protocol}

All models are evaluated with batch size 1 on a single NVIDIA
A100-SXM4-40GB GPU. We use the same ECGBench PTB-XL test example
(\texttt{ptb-xl-14149}) and the same benchmark question for all methods.
Image-based methods receive the rendered ECG image provided by ECGBench,
whereas signal-based methods receive a $12\times5000$ input corresponding
to a 10-second, 12-lead ECG sampled at 500\,Hz.

For latency measurements, we perform three warm-up iterations followed by
ten timed iterations and report the median. End-to-end generation uses one
warm-up and three timed iterations because of its substantially higher
cost. CUDA synchronization is performed within each timed region to ensure
that asynchronous GPU execution is included in the measurement.

To make autoregressive generation directly comparable, all models use
greedy decoding with
\texttt{min\_new\_tokens=max\_new\_tokens=256}.
Fixing the generated sequence length prevents differences in early
stopping behavior from being interpreted as computational differences.

We report the following quantities:
\begin{itemize}
    \item \textbf{Modality tokens}: the number of non-text tokens inserted
    into the LLM input to represent the ECG.
    \item \textbf{Prefill FLOPs}: analytical LLM forward-pass FLOPs required
    to process the complete input prompt before autoregressive decoding.
    \item \textbf{Prefill latency}: wall-clock latency of the LLM forward
    pass over the input prompt.
    \item \textbf{Time to first token (TTFT)}: latency of a complete
    generation call until the first generated token, including modality
    encoding, multimodal alignment, LLM prefill, and the first decoding
    step.
    \item \textbf{End-to-end latency}: wall-clock time required to generate
    all 256 output tokens.
\end{itemize}

TTFT and prefill latency measure different portions of the inference path.
For \method, GEM, and PULSE, prefill latency is measured after modality
embeddings have been computed and therefore isolates LLM prompt processing.
TTFT instead includes the complete model path and is consequently the
primary cross-model wall-clock measure of input-processing latency.
For ECG-R1, the public model implementation does not expose an equivalent
interface for independently timing the LLM prefill, so its reported
prefill latency additionally includes its modality encoders.

\subsection{FLOP Accounting}

We analytically compute the forward-pass FLOPs of the language model using
the convention that one multiply-accumulate operation corresponds to two
FLOPs. Let $p_{\mathrm{body}}$ denote the number of parameters in the
transformer blocks excluding the embedding table and language-model head,
$L$ the number of transformer layers, $d$ the hidden dimension, $V$ the
vocabulary size, $T$ the number of input tokens, and $n$ the number of
generated tokens. The prefill cost is
\begin{equation}
\mathrm{FLOPs}_{\mathrm{prefill}}
=
2p_{\mathrm{body}}T
+
4LdT^2
+
2dV,
\end{equation}
where the first term accounts for parameterized transformer operations and
the second accounts for the attention matrix products.

For autoregressive decoding with a KV cache, the corresponding cost for
$n$ generated tokens is \citep{hoffman, pope}
\begin{equation}
\mathrm{FLOPs}_{\mathrm{decode}}
=
2p_{\mathrm{body}}n
+
4Ld
\left(
nT+\frac{n(n-1)}{2}
\right)
+
2dVn.
\end{equation}
The $nT+n(n-1)/2$ term accounts for the increasing number of cached key
positions attended to across successive decoding steps. Embedding lookup
and element-wise operations such as normalization, rotary position
encoding, nonlinear activations, and softmax are excluded. We also exclude
the modality encoders from the FLOP calculation so that the reported
quantity isolates the computational consequence of the representation
supplied to the LLM. Their parameter counts and their contribution to
wall-clock TTFT are measured separately.

\subsection{Modality Tokenization and Timing Protocol}

Image tokenization differs across the evaluated models. GEM resizes the
input to $336\times336$ and therefore uses a fixed visual-token count. PULSE employs an AnyRes input strategy; for the evaluated PTB-XL images, the selected $672\times672$ configuration produces four high-resolution
crops in addition to the base image. ECG-R1 uses dynamic visual resolution with its released evaluation configuration capped at 768 visual tokens. In contrast, \method{} directly consumes the ECG waveform,
and its 111 modality tokens are fixed by the dimensions of the lead-temporal interface.

Finally, each latency result corresponds to the median of repeated timed
iterations for one benchmark example. Modality-token counts and analytical
FLOPs are deterministic for the evaluated input configuration, while the
observed prefill and TTFT timing ranges were below 2\% across repeated
iterations.

\subsection{Detailed Efficiency Results}

\begin{table}[t]
\centering
\caption{
Detailed inference-cost comparison. All latency measurements use batch size
1 on an NVIDIA A100-SXM4-40GB. FLOPs correspond to LLM computation;
end-to-end measurements generate exactly 256 tokens.
}
\label{tab:inference_efficiency_full}
\scriptsize
\setlength{\tabcolsep}{3.2pt}
\resizebox{\linewidth}{!}{
\begin{tabular}{lrrrrrrrrrr}
\toprule
\textbf{Method}
& \textbf{Non-LLM}
& \textbf{Modality}
& \textbf{Prefill}
& \textbf{Prefill}
& \textbf{Decode}
& \textbf{Total}
& \textbf{Prefill}
& \textbf{TTFT}
& \textbf{E2E}
& \textbf{Peak VRAM} \\
& \textbf{Params (M)}
& \textbf{Tokens}
& \textbf{Tokens}
& \textbf{FLOPs (T)}
& \textbf{FLOPs (T)}
& \textbf{FLOPs (T)}
& \textbf{(ms)}
& \textbf{(ms)}
& \textbf{(ms)}
& \textbf{(MB)} \\
\midrule
HeartLLM$^\dagger$
& \textbf{0.03}$^\dagger$ & 313 & 457 & \textbf{2.74}$^\dagger$ & \textbf{1.75}$^\dagger$ & \textbf{4.49}$^\dagger$
& 108.0 & 138.0 & 21,896 & \textbf{2,468} \\

GEM
& 411 & 676 & 802 & 10.73 & 3.51 & 14.23
& 65.8 & 90.7 & 8,678 & 14,787 \\

PULSE
& 324 & 2,879 & 3,005 & 43.66 & 3.80 & 47.46
& 242.7 & 270.9 & 8,972 & 16,993 \\

ECG-R1
& 682 & 862 & 936 & 13.52 & 4.04 & 17.56
& 145.9 & 149.1 & 14,200 & 17,389 \\

\midrule
\textbf{\method}
& 235 & \textbf{111} & \textbf{239}
& 3.13 & 3.43 & 6.56
& \textbf{33.1} & \textbf{43.6} & \textbf{8,344}
& 13,585 \\
\bottomrule
\end{tabular}
}

\raggedright
\footnotesize
$^\dagger$HeartLLM uses a 3.3B LLM quantized to 4-bit NF4 with unmerged
LoRA adapters, whereas the remaining models use 6.7--8.2B LLMs in
FP16/BF16. Its FLOPs and VRAM therefore reflect a different model size and
numeric execution path and should not be interpreted as a like-for-like
arithmetic comparison.
\end{table}

Table~\ref{tab:inference_efficiency_full} provides the full breakdown of
inference cost across the evaluated ECG--LLMs. The main differences arise
during input processing: \method{} uses the fewest modality tokens and
correspondingly has the lowest prefill latency and TTFT among the
full-precision models. In contrast, the end-to-end latency of \method{},
GEM, and PULSE is relatively similar despite large differences in
modality-token count, since the fixed 256-token output introduces a
substantial autoregressive decoding cost that is shared across these
similarly sized language models.

HeartLLM exhibits a different efficiency profile. Its quantized 3.3B
backbone results in substantially lower peak memory and analytical FLOPs,
but its measured latency is considerably higher. This illustrates that
analytical FLOPs are most informative when comparing models with similar
model sizes and numerical execution paths, whereas wall-clock latency can
also depend strongly on quantization and implementation details.

\color{black}
\subsection{Efficiency for Multiple Questions per ECG}
\label{app:multi_question_efficiency}

The single-question measurements do not fully capture the cost of answering multiple questions about the same recording. Question-independent methods can reuse their ECG representations across questions, whereas \method{} must recompute its question-conditioned interface outputs. We therefore evaluate how inference cost changes with the number of questions per ECG.

\paragraph{Evaluation protocol.}
We evaluate $K \in \{1,2,4,8,16,32\}$ questions per recording and report the cumulative prefill FLOPs and latency required to process all $K$ questions. Each model is allowed to reuse the KV cache for prompt content shared across questions. For question-independent methods, the ECG representation remains unchanged and can form part of the reusable prefix. For \method{}, the underlying ECG encoder features are question-independent and can be reused, but the instruction contextualizer and dual-attention interface must be evaluated for each question. Because the resulting ECG tokens change with the question, their corresponding LLM KV states must also be recomputed.

These experiments are conducted on an NVIDIA L4 GPU, whereas the single-question profiling experiments use an NVIDIA A100. Absolute latency values should therefore be compared only within the same experiment.
\begin{figure}[h]
    \centering
    \includegraphics[width=0.75\linewidth]{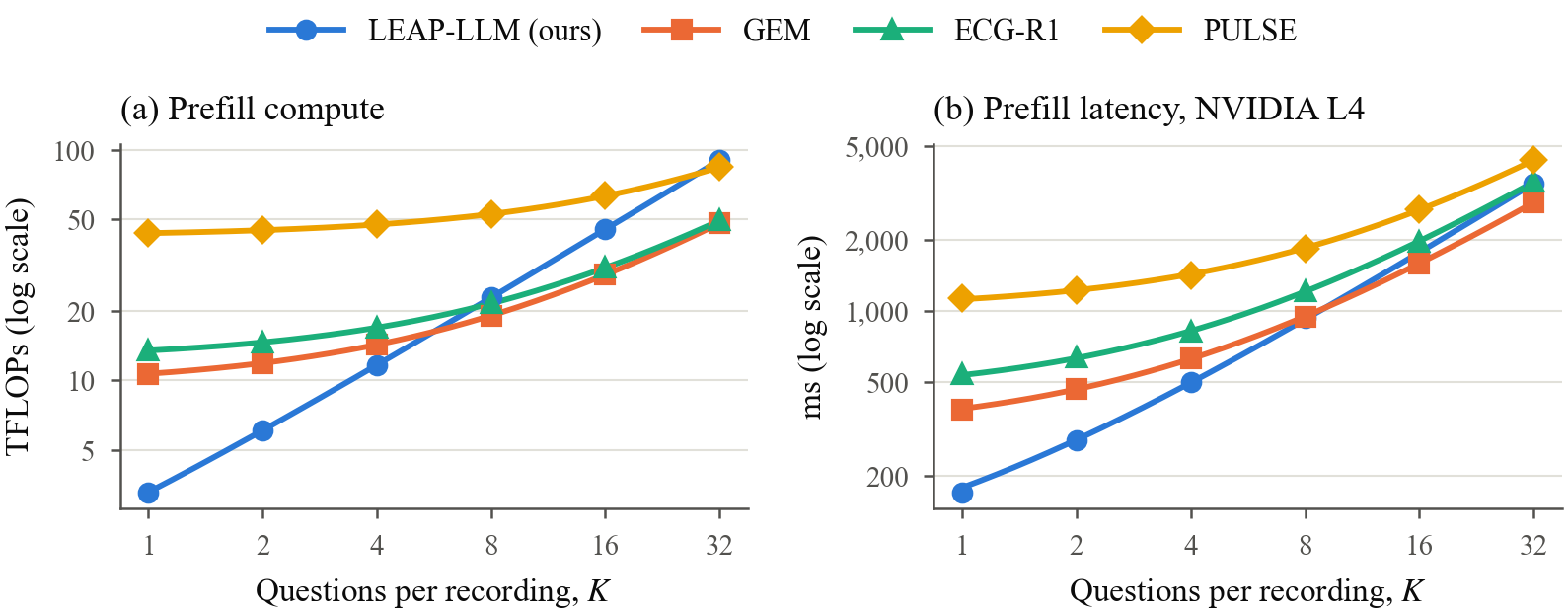}
    \caption{\textbf{Inference cost for multiple questions per ECG.}
    Cumulative prefill FLOPs (left) and latency (right) as the number of
    questions per recording increases.}
    \label{fig:multi_question_efficiency}
\end{figure}
\paragraph{Results.}
Figure~\ref{fig:multi_question_efficiency} shows how cumulative inference
cost scales with the number of questions per ECG. \method{} achieves the lowest cumulative prefill FLOPs among the evaluated models for $K=1,2,4$ and the lowest cumulative latency for $K=1,2,4,8$. Thus, its compact 111-token interface remains advantageous when answering a small number of questions about the same recording, despite requiring question-dependent extraction.

As the number of questions increases, GEM benefits more from reusing its question-independent ECG prefix. Among the evaluated question counts, GEM overtakes \method{} in cumulative prefill FLOPs at $K=8$ and in cumulative latency at $K=16$. Nevertheless, \method{} remains faster than PULSE and ECG-R1 at every evaluated question count, including $K=32$.

\color{black}
\section{Lead-Time Structural Analysis}
\label{app:attention_analysis}

We analyze the attention maps produced by the Temporal Branch to examine
whether the learned ECG-LLM interface organizes information jointly according
to lead identity and cardiac phase. All analyses use the PTB-XL test split from
ECG-Bench and the same fixed classification prompt, such that only the ECG
input varies across examinations. We analyze the raw post-softmax attention
weights without additional normalization or post-hoc z-scoring.

Because attention weights indicate where the interface concentrates
information but do not establish causal importance, we interpret this analysis
as evidence of model selectivity. Functional dependence on the resulting
lead representations is evaluated separately through the lead-token
interventions in Appendix~\ref{app:lead_lesion}.

\textbf{Myocardial infarction cohorts.}
We construct myocardial infarction cohorts from the PTB-XL SCP diagnostic
codes using an any-code inclusion criterion. Inferior myocardial infarction
includes \{IMI, ILMI, IPMI, IPLMI\}, with the clinically associated leads
II, III, and aVF, while anterior myocardial infarction includes
\{AMI, ASMI\}, associated with leads V1-V4~\citep{localmi}. This yields
310 inferior-MI examinations, 264 anterior-MI examinations, and 925 NORM
examinations. We use the NORM cohort as a common reference to determine
whether the lead-time attention patterns observed in each infarction group
differ from those present in normal ECGs.

\textbf{R-peak detection and cardiac-phase alignment.}
To align attention patterns across ECGs with different heart rates, we define
temporal position relative to the cardiac cycle. R peaks are detected from
lead II using the WFDB XQRS \citep{beatmatchalgo} algorithm. Each ECG patch
is assigned a temporal offset according to the distance between its center and
the nearest detected R peak. We retain ECGs with at least three detected
R peaks and at least one RR interval within the physiologically plausible range
of 0.3--2.0\,s (30--200\,bpm), which helps exclude records affected by
obvious detection errors. Using these criteria, 1,997 of the 2,082 PTB-XL test
ECGs (95.9\%) are successfully aligned for the phase-resolved analysis.

\textbf{Phase-resolved lead attention.}
The Temporal Branch operates independently at each temporal patch, with the
question-conditioned instruction tokens attending across the 12 ECG leads.
Thus, every temporal position is associated with a distribution of attention
across leads. Retaining these weights over all temporal patches produces a
two-dimensional lead--time attention map for each ECG.

Patch centers are grouped into 50\,ms phase bins according to their temporal
offset from the nearest R peak. Within each phase bin, we average the Temporal
Branch attention assigned to each of the 12 leads, yielding a phase-resolved
lead-attention map for every ECG. We denote the resulting attention assigned to
lead $c$ in cardiac-phase bin $\tau$ for ECG $x$ as $A_{c,\tau}(x)$.

To isolate disease-associated changes from global lead preferences learned by
the model, we compare each infarction cohort against the NORM reference rather
than interpreting absolute attention values. For each lead and cardiac-phase
bin, we compute
\begin{equation}
\Delta A_{c,\tau}
=
\mathbb{E}_{x\sim\mathcal{D}_{\mathrm{MI}}}
    [A_{c,\tau}(x)]
-
\mathbb{E}_{x\sim\mathcal{D}_{\mathrm{NORM}}^{\,w}}
    [A_{c,\tau}(x)],
\label{eq:lead_phase_contrast}
\end{equation}
where $\mathcal{D}_{\mathrm{MI}}$ denotes the corresponding inferior- or
anterior-MI cohort and $\mathcal{D}_{\mathrm{NORM}}^{\,w}$ denotes its
heart-rate-matched NORM reference cohort.

\paragraph{Uncertainty estimation.}
We estimate 95\% confidence intervals using 2,000 record-level bootstrap
resamples. Infarction and NORM examinations are independently resampled
within each comparison, and the heart-rate-matching weights and
lead-by-phase contrasts are recomputed for each bootstrap sample. The shaded
regions in Figure~\ref{fig:lead_phase_attention} denote the resulting
percentile confidence intervals.

\color{black}
\section{Question-Conditioned Lead Selection}
\label{app:question_swap}

We evaluate whether changing the question changes the lead evidence selected from the same ECG. Unlike the preceding analysis, which uses a fixed classification prompt across recordings, this experiment holds each ECG fixed and varies the queried myocardial infarction (MI) territory.

\textbf{Evaluation cohort.}
We use 2{,}082 held-out ECGs from PTB-XL fold 10. Based on their SCP diagnostic codes, the recordings are divided into six groups, as shown in Table~\ref{tab:question_swap_cohort}. The cohort includes normal ECGs, abnormal ECGs without MI, and ECGs with single- or multiple-territory MI.

\begin{table}[h]
\centering
\small
\caption{PTB-XL cohort used for the question-swap analysis.}
\label{tab:question_swap_cohort}
\begin{tabular}{lr}
\toprule
Group & ECGs \\
\midrule
Normal                  & 925 \\
Abnormal, without MI    & 644 \\
Inferior MI             & 224 \\
Anterior MI             & 160 \\
Lateral MI              & 23 \\
Multiple-territory MI   & 106 \\
\midrule
Total                   & 2{,}082 \\
\bottomrule
\end{tabular}
\end{table}

\textbf{Question-swap protocol.}
Each recording is paired with questions about inferior, anterior, and lateral MI, normal versus abnormal status, and the presence of any MI. For the territory-swap comparison, we examine the inferior-MI and anterior-MI questions while keeping the ECG unchanged. We compare the attention assigned to the inferior leads (II, III, and aVF) and anterior leads (V1--V4) under the two questions.

\textbf{Results.}
Table~\ref{tab:question_swap_attention} shows that lead attention changes substantially with the queried territory. The inferior-lead attention share is 79.8\% under the inferior-MI question but falls to 9.1\% under the anterior-MI question. Conversely, the anterior-lead share rises from 7.7\% under the inferior-MI question to 78.4\% under the anterior-MI question. Thus, the same ECG produces different lead-attention distributions depending on the question.

\begin{table}[h]
\centering
\small
\caption{Attention share (\%) assigned to each lead group when the queried MI territory changes while the ECG is held fixed.}
\label{tab:question_swap_attention}
\begin{tabular}{lcc}
\toprule
Question About & Inferior leads & Anterior leads \\
\midrule
Inferior MI & \textbf{79.8} & 7.7 \\
Anterior MI & 9.1 & \textbf{78.4} \\
\bottomrule
\end{tabular}
\end{table}

\begin{figure}[h]
\centering
\includegraphics[width=0.75\linewidth]{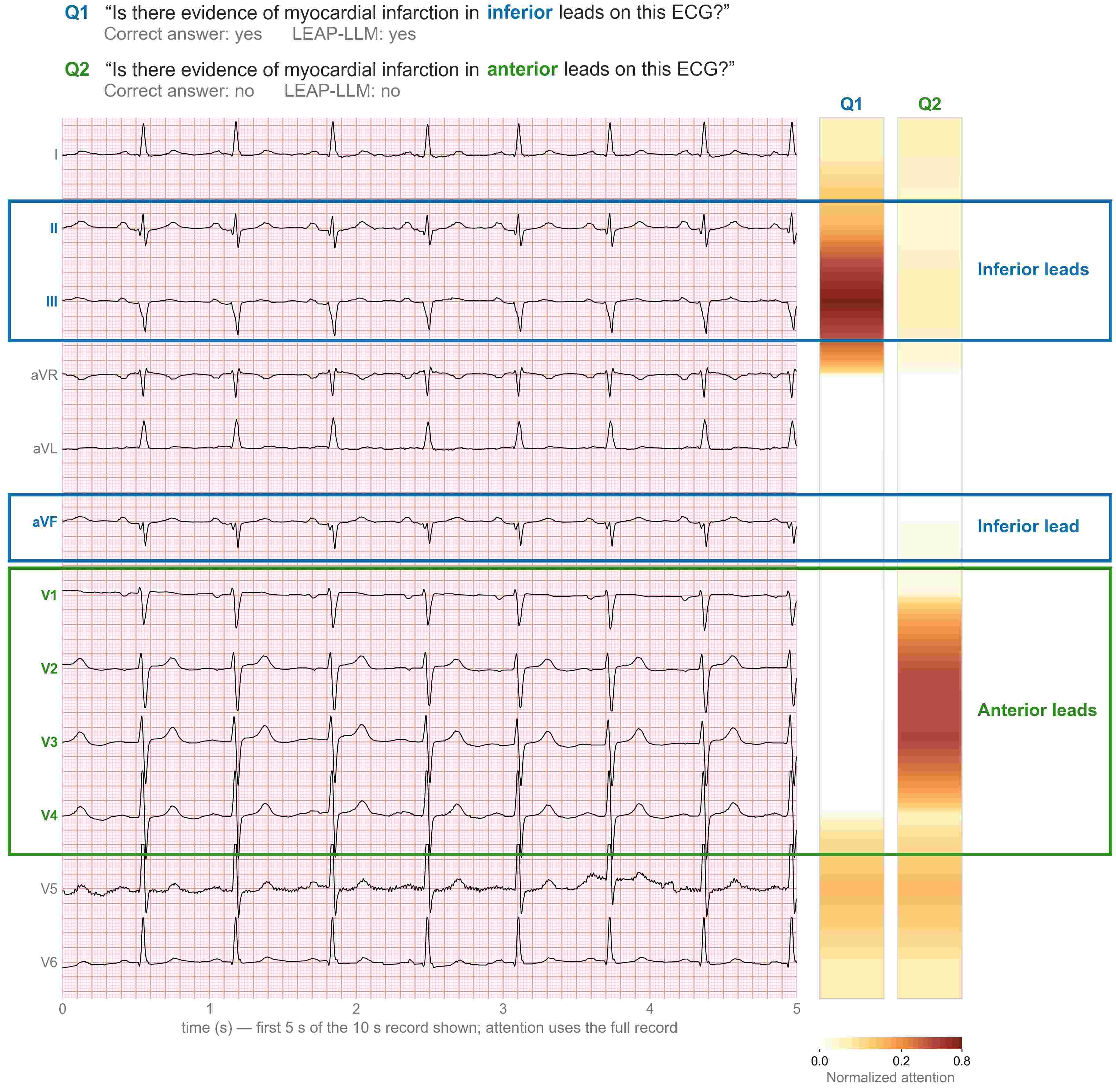}
\caption{Question-dependent lead attention for PTB-XL recording 6925. Changing the queried MI territory shifts attention toward the corresponding lead group while the ECG remains unchanged.}
\label{fig:question_swap_example}
\end{figure}

\paragraph{Example visualization.}
Figure~\ref{fig:question_swap_example} illustrates question-dependent lead selection for PTB-XL recording 6925. With the waveform unchanged, the attention distribution shifts toward the lead group associated with the queried territory.

These results demonstrate that the extraction module responds to the question rather than assigning a fixed lead-attention distribution to each recording. Attention alone does not establish that the selected representations affect the answer; this is examined separately through the lead-token interventions in Appendix~\ref{app:lead_lesion}.
\color{black}
\section{Functional Dependence on Lead-Specific Representations}
\label{app:lead_lesion}

\textbf{Lead-token intervention.}
The Lead Branch preserves one representation for each physical ECG lead. This allows individual lead tokens to be manipulated immediately before they are supplied to the LLM. For an intervened lead set $S$, we replace each patient-specific lead token with the mean representation of the corresponding lead computed over the reference cohort:

\begin{equation}
\widetilde{\mL}_c =
\begin{cases}
\overline{\mL}_c, & c \in S,\\
\mL_c, & c \notin S,
\end{cases}
\end{equation}

where $\mL_c$ is the original representation for lead $c$ and $\overline{\mL}_c$ is its dataset-level mean representation. Replacement is performed separately for each lead, thereby removing examination-specific information while preserving lead identity and avoiding the out-of-distribution perturbation introduced by zeroing a token. The temporal tokens and all non-intervened lead tokens remain unchanged.

We evaluate three primary interventions: replacing the inferior leads $\{\mathrm{II},\mathrm{III},\mathrm{aVF}\}$, replacing the anterior leads $\{\mathrm{V1},\ldots,\mathrm{V4}\}$, and replacing all 12 lead tokens.

\textbf{Likelihood-based evaluation.}
We quantify the effect of each intervention using the likelihood of the
ground-truth answer. For examination $i$ with target answer
$y_i=(y_{i,1},\ldots,y_{i,m_i})$, we first compute the mean token
log-likelihood
\begin{equation}
\ell_i =
\frac{1}{m_i}
\sum_{j=1}^{m_i}
\log p
\left(
y_{i,j}
\mid
y_{i,<j},\,\mathrm{ECG}_i,\,q_i
\right).
\end{equation}
For diagnostic group $G$ and intervention $S$, we first compute the
geometric-mean likelihood ratio between the intervened and intact
representations,
\begin{equation}
R(G,S)
=
\exp
\left(
\frac{1}{|G|}
\sum_{i\in G}
\left[
\ell_i^{S}
-
\ell_i^{\mathrm{intact}}
\right]
\right).
\end{equation}
We report the corresponding reduction in gold-answer likelihood,
\begin{equation}
D(G,S) = 1 - R(G,S),
\end{equation}
where $\ell_i^{S}$ and $\ell_i^{\mathrm{intact}}$ denote the mean token log-likelihoods under the intervened and original representations, respectively. A value of $D=0$ indicates no change in the model's support for the correct answer, while larger positive values indicate a greater reduction in likelihood and therefore stronger dependence on the replaced
representations. For ease of interpretation, we report $D(G,S)$ as the percentage
reduction in gold-answer likelihood.

\textbf{Control interventions.}
We include two controls to distinguish territory-specific dependence from generic sensitivity to token perturbation. First, we replace three randomly selected lead tokens, matching the number of tokens in the inferior-lead intervention. Secondly, we replace 12 randomly selected temporal tokens, matching the number of tokens removed by the all-lead intervention.

\textbf{Uncertainty estimation.}
We estimate 95\% confidence intervals \citep{bootstrap} using 2,000
record-level bootstrap resamples within each diagnostic group. For each
resample, we recompute the mean difference between the intervened and
intact log-likelihoods, convert it to the likelihood ratio $R$, and report
the corresponding likelihood reduction $D=1-R$. Confidence intervals are
obtained from the bootstrap distribution of $D$.
\begin{table}[h]
\centering
\caption{
Full lead-token intervention results. Values report the reduction in
gold-answer likelihood, $D=1-R$, with 95\% bootstrap confidence intervals.
Higher values indicate greater dependence on the replaced representations.
Bold values denote the anatomically matched territory-specific interventions.
}
\label{tab:lead_lesion_app}
\small
\setlength{\tabcolsep}{10pt}
\renewcommand{\arraystretch}{1.15}

\begin{tabular}{lcc}
\toprule
\textbf{Tokens replaced}
& \textbf{Inferior MI}
& \textbf{Anterior MI} \\
\midrule

All leads
& \shortstack{73.5\% \\ {\scriptsize [71.3, 75.5]}}
& \shortstack{80.7\% \\ {\scriptsize [78.3, 82.7]}} \\

Inferior leads
& \shortstack{\text{50.5\%} \\ {\scriptsize \text{[46.1, 54.4]}}}
& \shortstack{5.5\% \\ {\scriptsize [1.6, 9.4]}} \\

Anterior leads
& \shortstack{6.8\% \\ {\scriptsize [3.2, 11.0]}}
& \shortstack{\text{55.0\%} \\ {\scriptsize \text{[49.9, 59.5]}}} \\

Random three leads
& \shortstack{17.2\% \\ {\scriptsize [13.6, 20.8]}}
& \shortstack{17.6\% \\ {\scriptsize [11.8, 23.4]}} \\

Random 12 temporal tokens
& \shortstack{$-0.2$\% \\ {\scriptsize [$-0.5$, 0.1]}}
& \shortstack{$-0.2$\% \\ {\scriptsize [$-0.6$, 0.1]}} \\

\bottomrule
\end{tabular}
\end{table}

\textbf{Additional results.}
The control interventions further support the territory specificity of the
main findings. Replacing three randomly selected lead tokens reduces the
gold-answer likelihood by 17.2\% for inferior MI and 17.6\% for anterior
MI, substantially smaller than the 50.5\% and 55.0\% reductions produced
by replacing the anatomically matched lead groups, respectively. In
contrast, replacing 12 randomly selected temporal tokens has essentially
no effect, with $D=-0.2\%$ for both infarction groups.

\section{Limitations and Future Work}

\paragraph{Limitations.}
Our evaluation is limited to standardized 10-second, 12-lead ECG recordings sampled at 500\,Hz, and therefore does not establish how \method{} behaves under different recording durations, sampling rates, or reduced-lead configurations. In addition, our current experiments use a limited set of LLM backbones, we do not systematically evaluate \method{} across a broader range of LLMs , so the generality of the proposed interface across architectures remains to be further established. Finally, our interpretability analysis is limited to inferior and anterior myocardial infarction, and therefore does not yet establish whether the same lead-selective behavior holds across other cardiac abnormalities.

\paragraph{Future Work.}
Future work will extend \method{} to more diverse ECG acquisition settings, including variable recording durations, sampling rates, and reduced- or missing-lead configurations. We also plan to evaluate the proposed interface across a broader range of LLM backbones to better characterize its architectural generality. Finally, we will expand the interpretability analysis beyond inferior and anterior myocardial infarction to additional rhythms, conduction abnormalities, and morphological findings, providing a more comprehensive view of how \method{} uses lead-specific and temporal information across diagnostic settings.

\end{document}